%% file: main.tex
\documentclass[10pt,twocolumn,letterpaper]{article}

\usepackage{cvpr}              % To produce the CAMERA-READY version
\input{preamble}

\definecolor{cvprblue}{rgb}{0.21,0.49,0.74}
\usepackage[pagebackref,breaklinks,colorlinks,allcolors=cvprblue]{hyperref}

\def\paperID{9167} % *** Enter the Paper ID here
\def\confName{CVPR}
\def\confYear{2026}

\title{Generative Digital Twins: \\Vision-Language Simulation Models for Executable Industrial Systems}

\author{YuChe Hsu \quad AnJui Wang \quad TsaiChing Ni \quad YuanFu Yang\\
Institute of Artificial Intelligence Innovation, National Yang Ming Chiao Tung University\\
{\tt\small danielhsu.ii13@nycu.edu.tw \quad anguswang.ii14@nycu.edu.tw}\\
{\tt\small nina.ii13@nycu.edu.tw \quad yfyangd@gmail.com}
}

\begin{document}
\maketitle

\begin{strip}
\centering
\vspace*{-20pt}
\includegraphics[width=\textwidth]{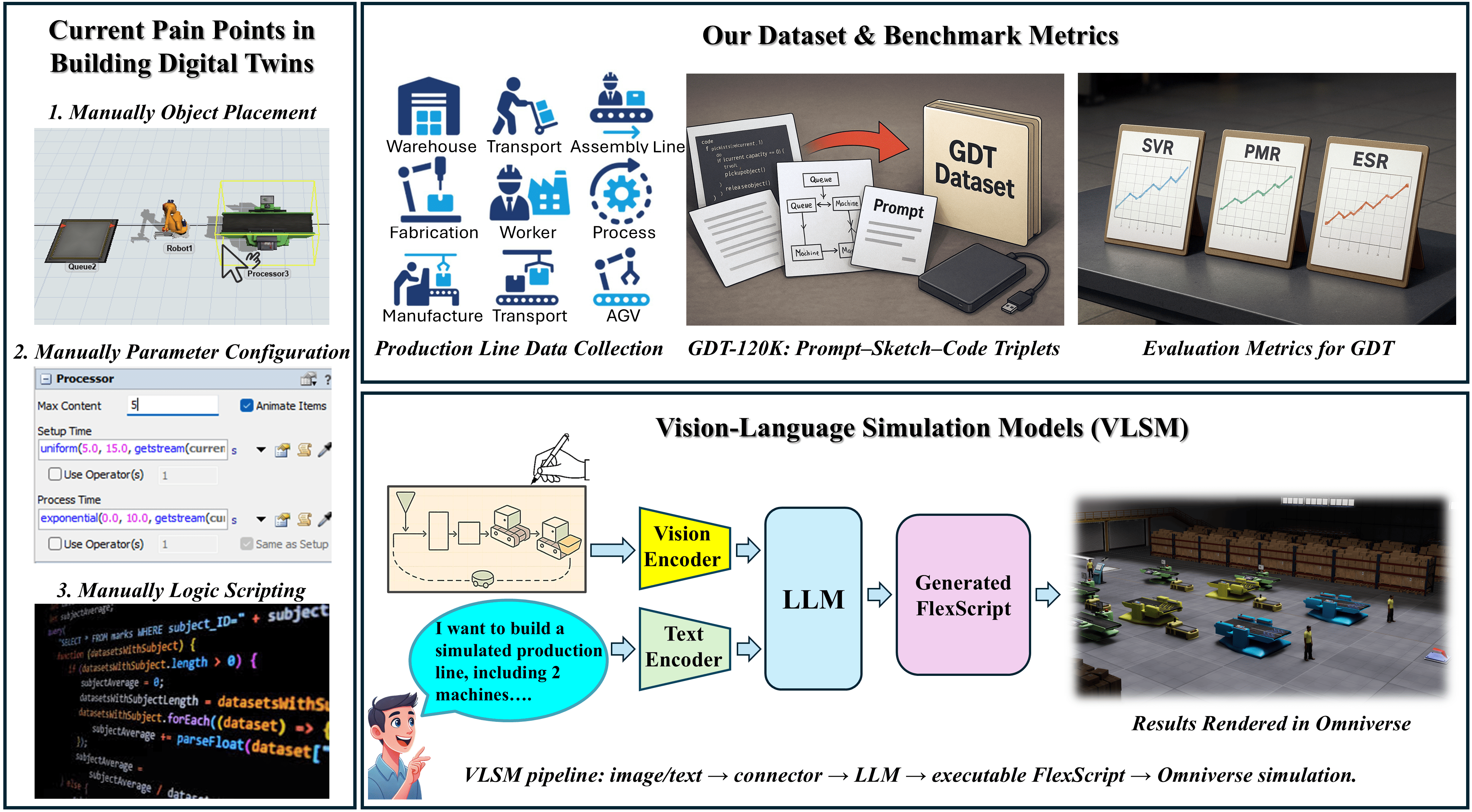}%
{\captionsetup{ hypcap=false, format=plain, justification=raggedright, singlelinecheck=false, font=small } 
\captionof{figure}{Overview of the generative digital twins framework, showing major challenges in digital-twin modeling, the construction of the GDT-120K dataset with evaluation metrics, and the Vision-Language Simulation Models (VLSM) workflow.} \label{fig:first_pic} }
\vspace*{-6pt}
\end{strip}

\input{sec/0_abstract}

\begin{figure*}[t]
    \centering
    \includegraphics[width=\textwidth]{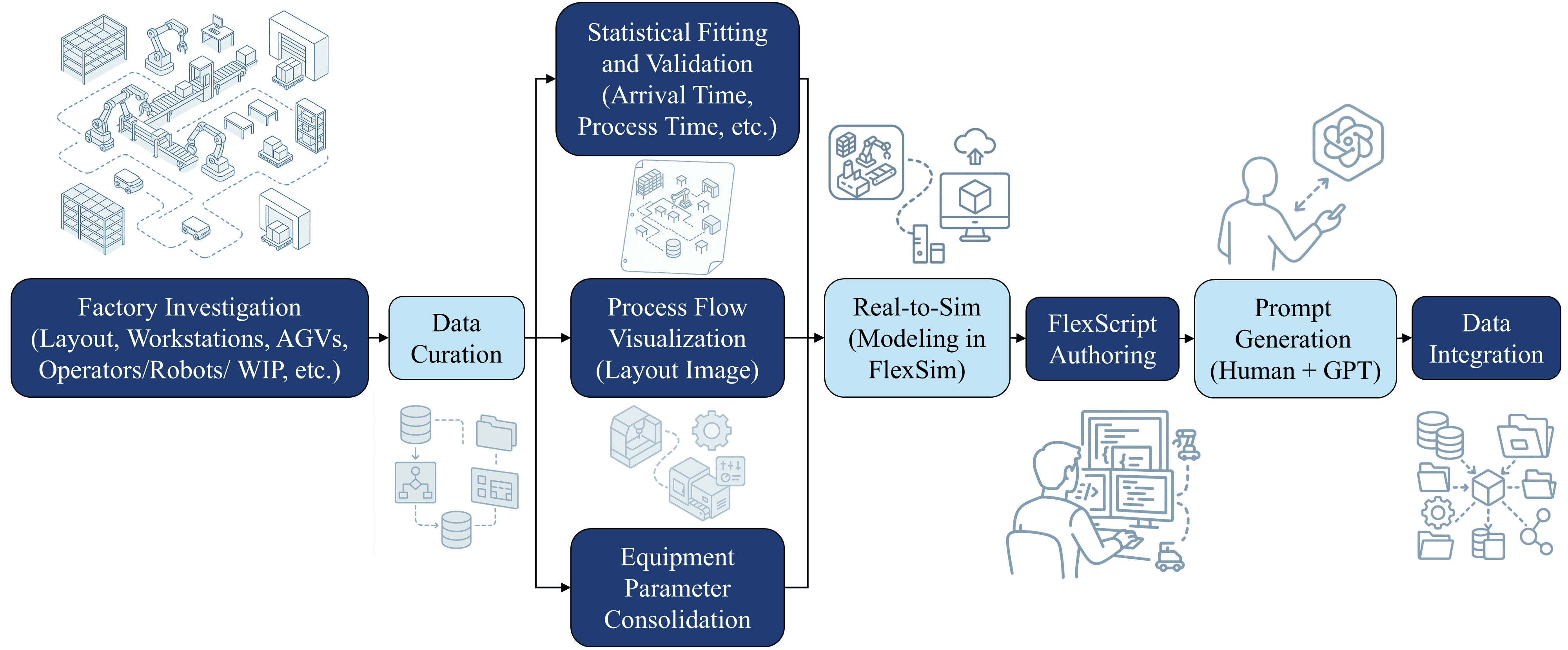}
    \caption{Workflow of the GDT-120K dataset construction, integrating curated factory data, statistical validation, and FlexSim instantiation with human–AI co-authored prompts to create aligned multimodal pairs for model training and evaluation.}
    \label{fig:dataset_pipeline}
\vspace*{-10pt}
\end{figure*}

\input{sec/1_intro}
\begin{figure*}[t]
    \centering
    
    \begin{subfigure}[t]{0.48\textwidth}
        \centering
        \includegraphics[width=\linewidth]{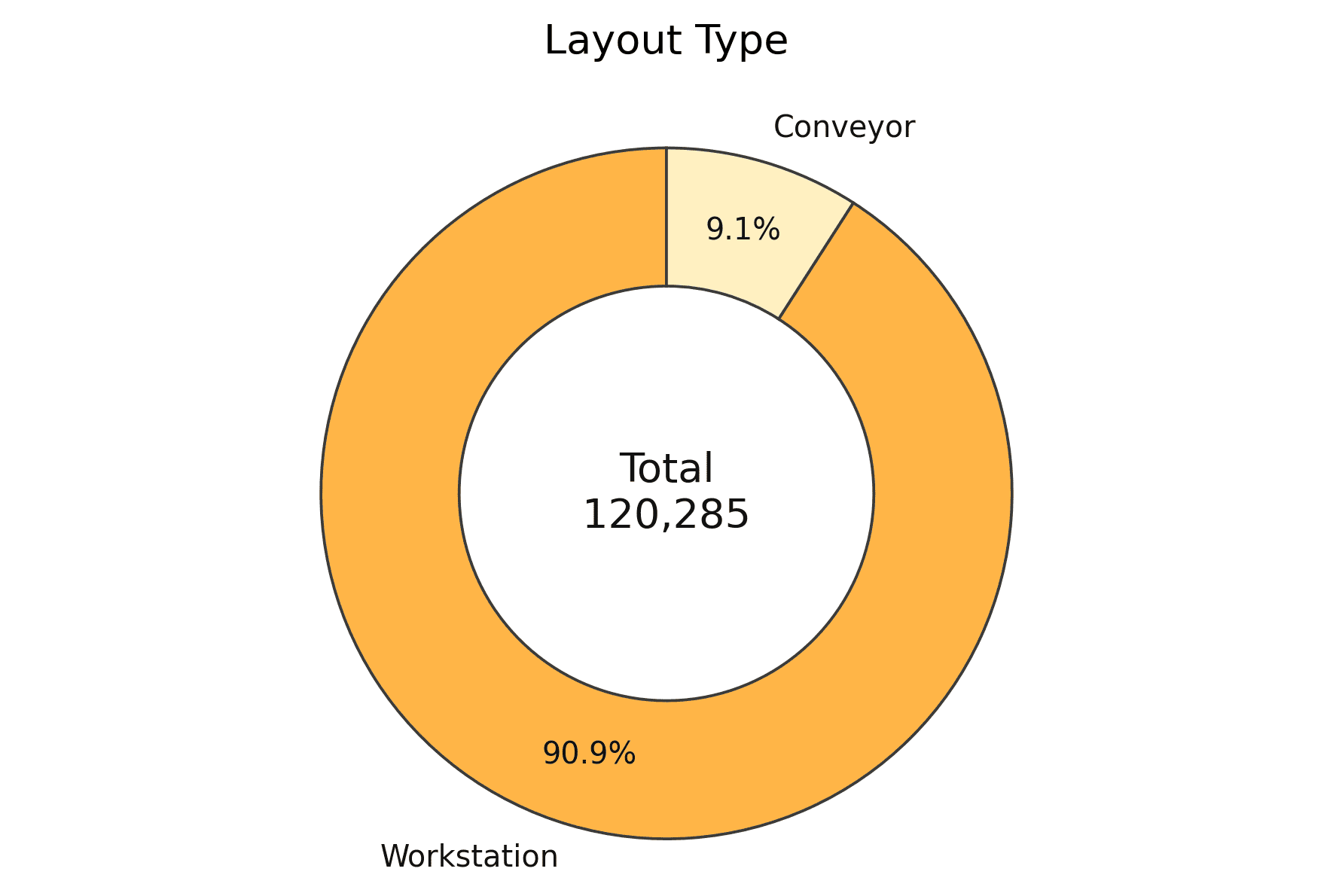}
        \caption{Layout type distribution}
        \label{fig:layout_type}
    \end{subfigure}\hfill
    \begin{subfigure}[t]{0.48\textwidth}
        \centering
        \includegraphics[width=\linewidth]{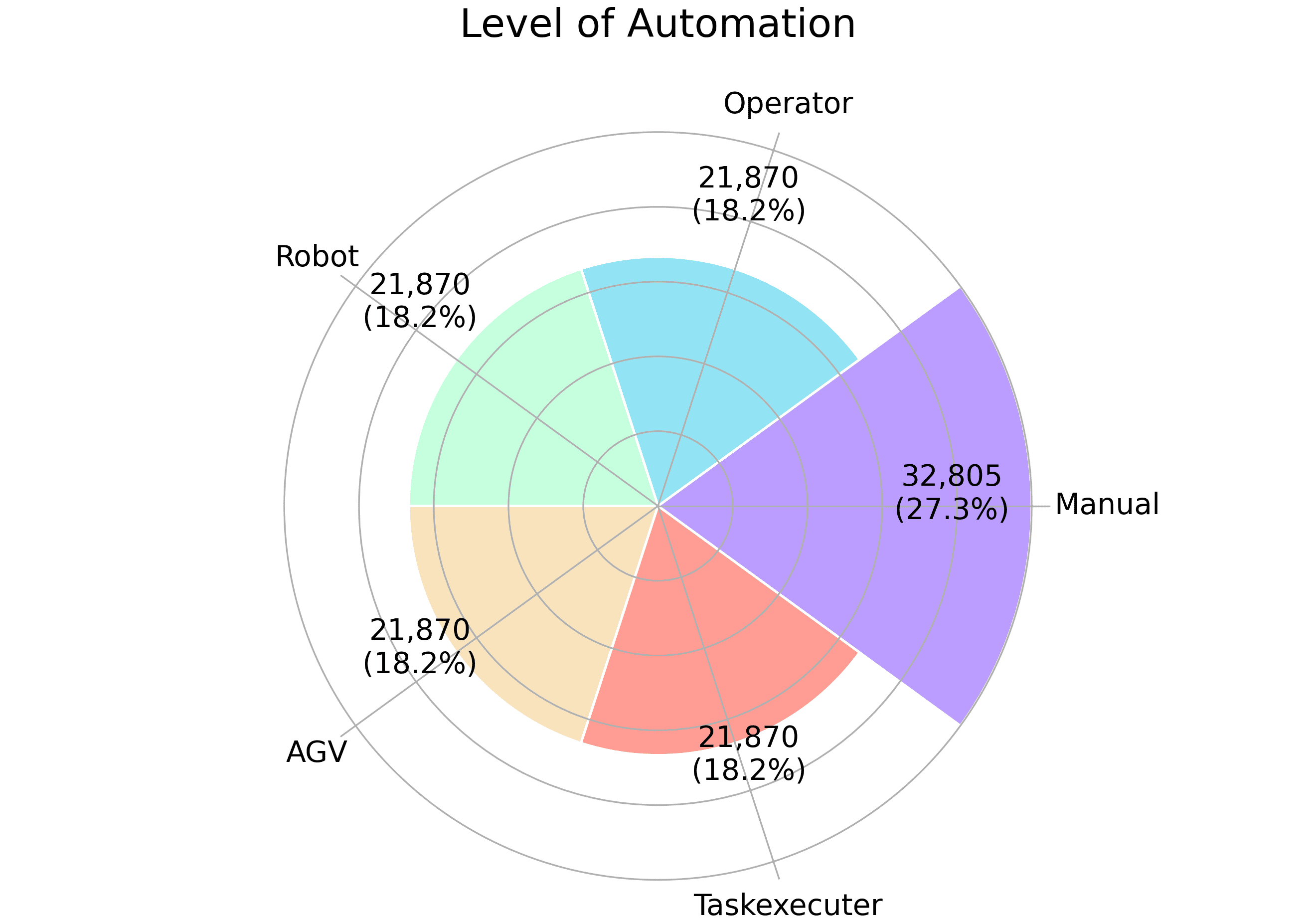}
        \caption{Level of automation}
        \label{fig:automation}
    \end{subfigure}

    \vspace{0.6em}

    \begin{subfigure}[t]{0.48\textwidth}
        \centering
        \includegraphics[width=\linewidth]{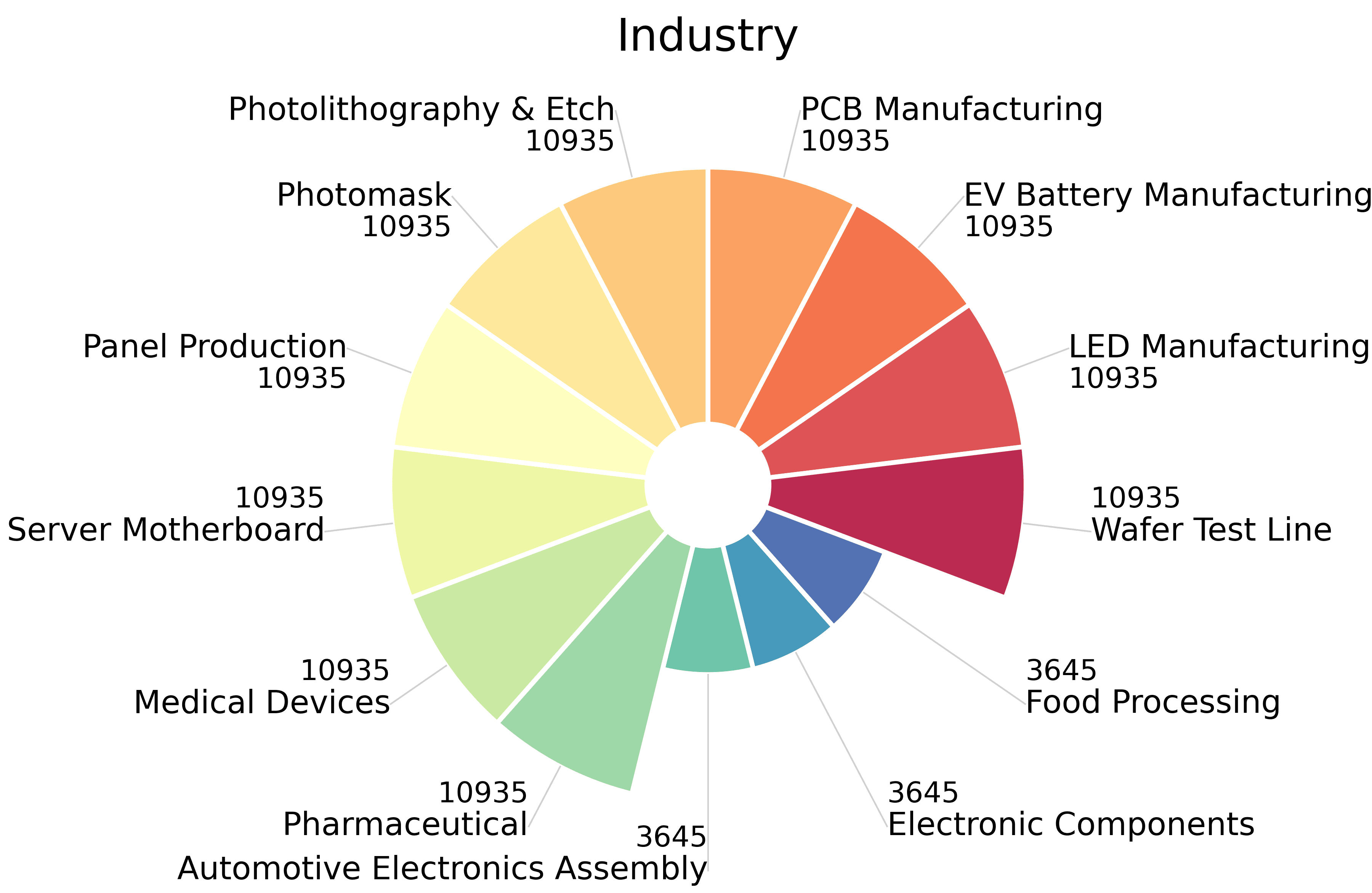}
        \caption{Industry coverage}
        \label{fig:industry}
    \end{subfigure}
    \begin{subfigure}[t]{0.48\textwidth}
        \centering
        \includegraphics[width=\linewidth]{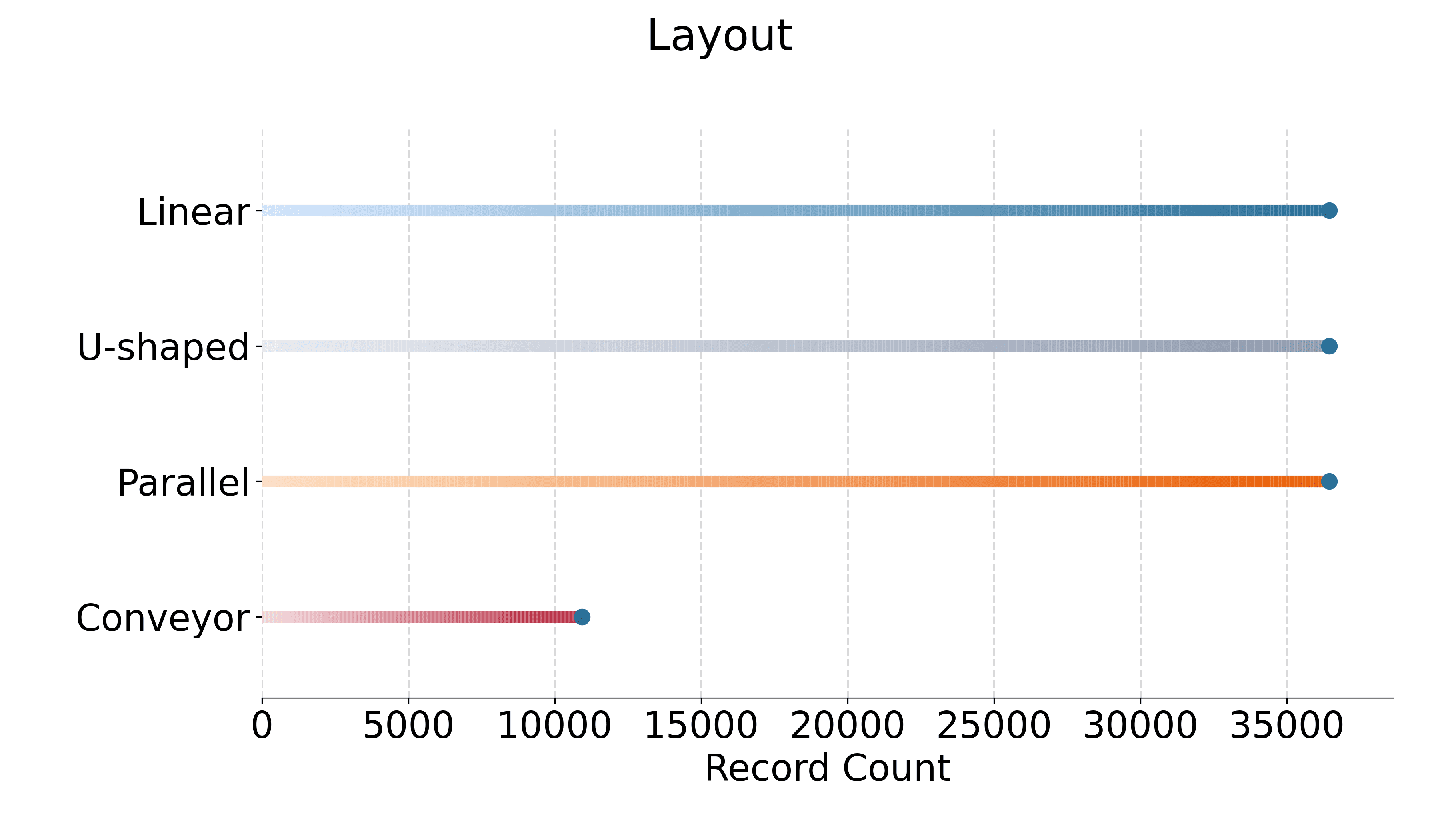}
        \caption{Layout categories}
        \label{fig:layout}
    \end{subfigure}\hfill

    \caption{Dataset statistics across layout type, automation level, industry and layout categories.}
    \label{fig:dataset_stats}
    \vspace*{-10pt}
\end{figure*}
\vspace*{-10pt}
\input{sec/2_related_work}

\input{sec/3_dataset} 
\input{sec/4_methdology}
\input{sec/5_experiments}

\input{sec/6_conclusion}
{
    \small
    \bibliographystyle{ieeenat_fullname}
    \bibliography{main}
}
\input{sec/X_suppl}

\end{document}

%% file: preamble.tex
\usepackage{tikz}
\usetikzlibrary{arrows.meta,positioning,fit,shapes.multipart}
\usepackage{listings}
\usepackage{xcolor}
\usepackage{array}
\usepackage{graphicx}
\usepackage{cuted}
\usepackage{capt-of}

%% file: sec/0_abstract.tex
\begin{abstract}
We propose a Vision-Language Simulation Model (VLSM) that unifies visual and textual understanding to synthesize executable FlexScript from layout sketches and natural-language prompts, enabling cross-modal reasoning for industrial simulation systems. To support this new paradigm, the study constructs the first large-scale dataset for generative digital twins, comprising over 120,000 prompt–sketch–code triplets that enable multimodal learning between textual descriptions, spatial structures, and simulation logic. In parallel, three novel evaluation metrics, Structural Validity Rate (SVR), Parameter Match Rate (PMR), and Execution Success Rate (ESR), are proposed specifically for this task to comprehensively evaluate structural integrity, parameter fidelity, and simulator executability. Through systematic ablation across vision encoders, connectors, and code-pretrained language backbones, the proposed models achieve near-perfect structural accuracy and high execution robustness. This work establishes a foundation for generative digital twins that integrate visual reasoning and language understanding into executable industrial simulation systems.
\end{abstract}

%% file: sec/1_intro.tex
\section{Introduction}
Digital twins replicate physical systems in virtual environments for process monitoring and optimization~\cite{barricelli2019survey, tao2019smart}. 
With increasing adoption in smart manufacturing~\cite{kritzinger2018digital}, platforms such as FlexSim are widely used for modeling complex production workflows. 
However, as shown in Figure~\ref{fig:first_pic}, building digital twins in FlexSim remains highly labor-intensive, requiring manual object placement, parameter configuration, and logic scripting in the proprietary FlexScript language. 
This process is time-consuming and difficult to scale, motivating a transition from manual authoring toward automated code synthesis.

Recent progress in large language models (LLMs)~\cite{openai2023gpt4, google2023gemini} suggests a pathway for automation. 
While LLMs can generate structured code from text and have been applied to control and planning~\cite{ahn2022saycan}, they lack visual grounding, which is crucial for spatially organized domains like factory layout synthesis. We therefore propose Vision-Language Simulation Models (VLSM), a multimodal generative framework that integrates visual and textual inputs to synthesize executable FlexScript for digital twins. 
Our goal is to allow users to specify a layout via sketches and prompts, enabling AI-assisted simulation authoring.

Operationalizing this paradigm is challenging due to the domain-specific nature of FlexScript, which diverges from general-purpose languages and lacks public datasets or suitable evaluation metrics. To address these gaps, we introduce the generative digital twins (GDT) framework, a large-scale multimodal dataset containing 120K prompt–sketch–code triplets together with three evaluation metrics, Structural Validity Rate (SVR), Parameter Match Rate (PMR), and Execution Success Rate (ESR). By integrating multimodal models that unify visual encoders and language backbones through joint training, the GDT framework provides a solid technical foundation for scalable and reliable digital twin generation.

%% file: sec/2_related_work.tex
\section{Related Work}
Models such as Codex~\cite{chen2021codex}, AlphaCode~\cite{li2022alphacode}, and StarCoder2~\cite{li2023starcoder} demonstrate strong general-purpose coding ability. For domain-specific DSLs (e.g., SQL, Verilog), specialized datasets and structure-aware training are required~\cite{liang2022cap, suris2023vipergpt}. In robotics and simulation, GenSim~\cite{wang2024gensim} and EnvGen~\cite{zala2024envgen} use LLMs to bootstrap synthetic scenarios, while MineDojo~\cite{fan2022minedojo} and Voyager~\cite{wang2023voyager} show agents writing environment scripts. These systems, however, do not target structured layout scripting as in FlexSim.

The integration of vision and language has been explored in image captioning~\cite{li2022blip}, VQA~\cite{lu2019vilbert}, and instruction following agents~\cite{liu2023llava, alayrac2022flamingo}. Recent LMMs (e.g., GPT-4~\cite{openai2023gpt4}, Kosmos-1~\cite{huang2023kosmos}) incorporate visual encoders into LLMs via modular fusion (e.g., Q-Former~\cite{li2023blip2}), allowing image-guided code or text generation. Unlike prior work that focuses on language or classification, our task requires precise layout-aware code generation.

While recent works have explored digital twins for monitoring and planning, few directly target full code synthesis for simulation engines. Existing approaches often rely on manual modeling or parameter tuning interfaces. In contrast, our work generates complete and executable simulation programs from scratch, conditioned only on natural inputs consisting of textual descriptions and layout sketches. To the best of our knowledge, this is the first multimodal system that translates high-level specifications into runnable digital-twin logic for industrial simulation platforms.

%% file: sec/3_dataset.tex
\section{GDT Dataset}
\subsection{Construction and Modalities}
\label{sec:dataset_construction}
Our workflow converts factory information into a multimodal dataset, as illustrated in Figure~\ref{fig:dataset_pipeline}. It begins with factory investigations collecting layout data, workstation types, and production resources. Statistical fitting is applied to validate and normalize timing attributes such as arrival and process times. In parallel, layout sketches are drawn to convey workstation order and queue placement, while equipment parameters are standardized into a canonical schema. The curated information is instantiated in FlexSim to generate reference projects, from which corresponding FlexScript is derived. Prompts are drafted from metadata with GPT assistance and refined by human editors, and a subset of samples includes paired layout sketches.

\subsection{Dataset Design and Realization}
\label{sec:data_expansion}

This subsection outlines the design principles and statistical structure underlying our dataset. We adopt a five-layer framework encompassing the production line process, parameter diversity, level of automation, industry type, and layout configuration. 

Guided by this framework, the dataset emulates real industrial projects while maintaining diversity across key generative and structural dimensions. The production line settings cover workstation- and conveyor-based layouts, as shown in Figure~\ref{fig:layout_type}, ensuring representative coverage of spatial and operational topologies. This design highlights the generality of the GDT framework over any specific simulation platform. The level of automation in Figure~\ref{fig:automation} covers manual, operator, robot, automated guided vehicle (AGV), and task-executor modes, ensuring coverage from manual handling to control-driven execution.

To enhance domain generalization, the dataset spans thirteen industries, balancing mainstream sectors such as semiconductor and electronics with specialized ones like photomask and food processing (as shown in Figure~\ref{fig:industry}). Layout configurations include linear, U-shaped, parallel, and conveyor forms (as shown in Figure~\ref{fig:layout}), expanding the spatial diversity and routing complexity encountered during training for more robust, topology-aware generation.

\begin{table}[t]
\centering
\caption{Arrival time distributions for \emph{sources}.}
\label{tab:src_dists}
\setlength{\tabcolsep}{2pt}
\resizebox{\columnwidth}{!}{%
\begin{tabular}{l l l}
\toprule
Distribution & Typical use & Parameterization \\
\midrule
Constant     & Fixed interarrival & \texttt{constant(c)} \\
Exponential  & Memoryless inflow  & \texttt{exponential(}$\lambda^{-1}$\texttt{)} \\
Normal       & Natural variation  & \texttt{normal(}$\mu, \sigma$\texttt{)} \\
Triangular   & Bounded with mode  & \texttt{triangular(}$a, m, b$\texttt{)} \\
Uniform      & Bounded unknown mode & \texttt{uniform(}$a, b$\texttt{)} \\
\bottomrule
\end{tabular}%
}
\end{table}

\begin{table}[t]
\centering
\caption{Service-time distributions for \emph{machines} (processor, separator, combiner, multiprocessor).}
\label{tab:mach_dists}
\setlength{\tabcolsep}{2pt}
\resizebox{\columnwidth}{!}{%
\begin{tabular}{l l l}
\toprule
Distribution & Typical use & Parameterization \\
\midrule
Constant     & Deterministic service     & \texttt{constant(c)} \\
Exponential  & Random short jobs         & \texttt{exponential(}$\lambda^{-1}$\texttt{)} \\
Normal       & Symmetric variability     & \texttt{normal(}$\mu, \sigma$\texttt{)} \\
Triangular   & Bounded with mode         & \texttt{triangular(}$a, m, b$\texttt{)} \\
Uniform      & Bounded without mode      & \texttt{uniform(}$a, b$\texttt{)} \\
Lognormal    & Skewed processing times   & \texttt{lognormal(}$\mu_{\ell}, \sigma_{\ell}$\texttt{)} \\
Weibull      & Reliability and wear      & \texttt{weibull(}$k, \lambda$\texttt{)} \\
Gamma        & Multi-stage effects       & \texttt{gamma(}$\alpha, \theta$\texttt{)} \\
Poisson      & Count-based servicing     & \texttt{poisson(}$\nu$\texttt{)} \\
\bottomrule
\end{tabular}%
}
\vspace*{-10pt}
\end{table}

Parameterization follows a canonical stochastic design that drives combinatorial diversity. Source arrivals are sampled from five distributions that capture typical inflow patterns (as shown in Table~\ref{tab:src_dists}), while machines adopt nine families that cover service time variability (as shown in Table~\ref{tab:mach_dists}). A production layout on average contains three machines, which induces 3645 (5 × 9 × 9 × 9) parameter combinations per layout. Sampling is performed within validated ranges per industry and automation level so that generated FlexScript remains executable and representative of realistic factory dynamics. These forms capture distinct spatial structures and routing constraints, and together with the other layers they anchor a corpus of 120{,}285 programs that reflect real-world variability.

%% file: sec/4_methdology.tex
\section{Methodology}
\label{sec:method}

\begin{figure*}[t]
    \centering
    \includegraphics[width=\textwidth]{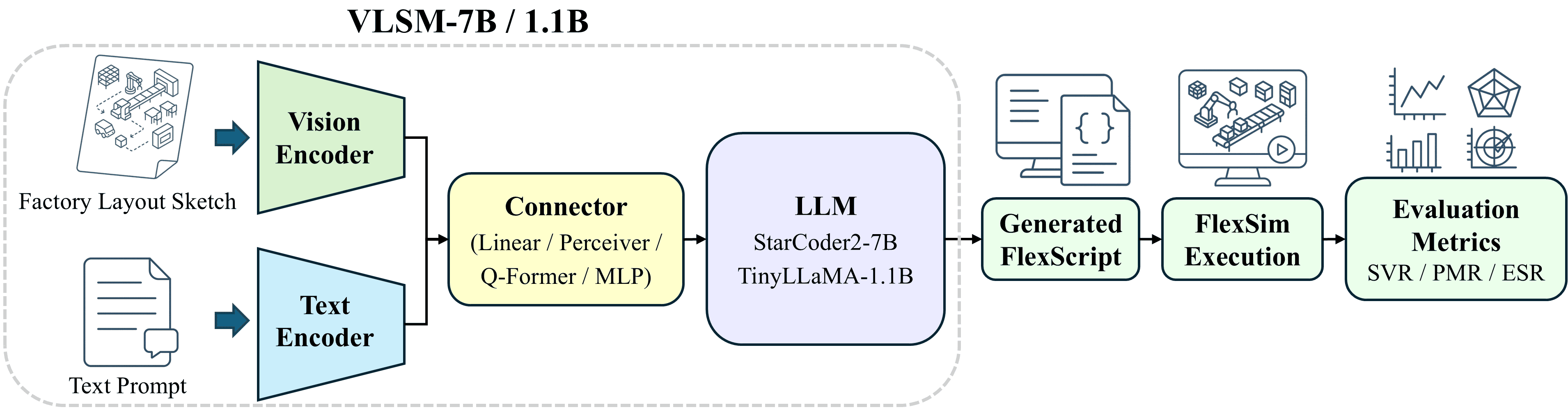}
    \caption{Overall architecture of the VLSM, integrating vision and language modules to generate executable FlexScript evaluated by SVR, PMR, and ESR (with BLEU as a supplementary metric).}
    \label{fig:overall_arch}
    \vspace*{-10pt}
\end{figure*}

\subsection{Overall Architecture}
\label{sec:overall}
To realize the GDT framework, we design a multimodal VLSM that integrates textual and visual inputs to generate executable FlexScript in FlexSim. As shown in Figure~\ref{fig:overall_arch}, VLSM supports both text-only and multimodal pathways. The vision encoder and trainable connector fuses layout features with language embeddings, and the decoded scripts are executed to evaluate SVR, PMR, and ESR for structural, parametric, and executable fidelity.

\subsection{LLM Backbone and Adaptation}
\label{sec:llm}
To support text-to-FlexScript generation, we evaluate seven open-source LLMs that balance deployment efficiency and code generation capability. We emphasize lightweight backbones to reduce inference cost and enable on-premises deployment for small and medium-sized enterprises with limited compute resources. Specifically, Gemma3-270M and TinyLLaMA-1.1B~\cite{gemma3_2025, tinyllama_2024} are fully retrained for domain specialization, while LLaMA2-7B, LLaMA3-8B, CodeLLaMA-7B, StarCoder2-7B, and Mistral-7B are fine-tuned with 4-bit QLoRA~\cite{touvron2023llama2,llama3_2024,roziere2023codellama,li2023starcoder,jiang2023mistral,dettmers2023qlora} to preserve strong reasoning and coding priors within commodity GPU memory.

\subsection{Visual Encoder for VLSM}
\label{sec:vlm}
To integrate spatial perception into FlexScript generation, the VLSM accepts a single image input such as a layout sketch, which encodes workstation order and queue placement. For the visual encoder, we adopt two CLIP-based model families. The first is OpenAI CLIP (ViT series)~\cite{radford2021clip}, trained on WebImageText and widely used in multimodal frameworks including LLaVA, MiniGPT-4, and TinyLLaVA~\cite{liu2023llava, zhu2023minigpt4, zhou2024tinyllava}. The second is LAION OpenCLIP (ViT-g/14), pre-trained in LAION-2B and serves as a strong frozen backbone for downstream vision-language tasks such as BLIP-2, InstructBLIP and Shikra~\cite{li2023blip2, dai2023instructblip, chen2023shikra}. The extracted visual features are subsequently fused with textual embeddings through the connector described in Section~\ref{sec:lmm}.

\subsection{Multimodal Fusion Module}
\label{sec:lmm}
The multimodal fusion module bridges visual representations from the encoder with linguistic embeddings from the language backbone, enabling coherent text-to-simulation generation. Built on a TinyLLaVA-style modular design, the module explores four connector types for visual–language alignment. A Linear Projection performs dimensional mapping between modalities. A Perceiver-style Resampler condenses visual tokens into compact latent arrays while preserving global structure. A Q-Former-style transformer extracts task-relevant information through learnable queries. A Two-Layer MLP introduces lightweight nonlinearity with minimal computational overhead. In combination with CLIP and OpenCLIP encoders, these connectors yield eight multimodal configurations per backbone. Section~\ref{sec:Ablation Study} presents a systematic ablation of their accuracy, latency, and stability, and the configuration with the best trade-off is adopted in our final model.

\subsection{Evaluation Metrics for GDT}
\label{sec:metrics}

Evaluating FlexScript generation requires metrics that reflect both structural and executable fidelity, as conventional text-based measures like BLEU-4~\cite{papineni2002bleu} fail to capture simulation accuracy. We therefore introduce three structural metrics tailored to simulation-oriented code generation, assessing syntactic validity, semantic correctness, and execution feasibility.
\\\textbf{Structural Validity Rate (SVR)} evaluates whether a generated script conforms to correct structural logic. We define the \emph{connection score (CS)} by extracting the set of \texttt{contextdragconnection(\dots)}~\cite{flexsim_commandgroups_24_2} statements and comparing them with the ground-truth. If $N$ denotes the total number of ground-truth connections and $M$ denotes the number reproduced correctly, the connection score is defined as follows:
\begin{equation}\label{eq:cs}
\mathrm{CS}=\frac{M}{N}.
\end{equation}
We further define the \emph{object score (OS)} by verifying that all connected objects are properly declared. The declared \texttt{objType} must be correct (for example, \texttt{/source}, \texttt{/queue}) and the \texttt{objName} must exactly match the ground-truth with case sensitivity. If $K$ denotes the total number of required objects and $K'$ denotes the number validly declared, the object score is defined as follows:
\begin{equation}\label{eq:os}
\mathrm{OS}=\frac{K'}{K}.
\end{equation}
The final structural validity is computed as a weighted combination, giving greater importance to the connection score since correct connectivity defines the process topology governing flow routing, blocking, and scheduling. An incorrect connection may reroute items, cause dead ends, or create unintended cycles, thereby invalidating execution even if object declarations remain correct. In contrast, declaration mismatches are typically local and recoverable. The final SVR is thus defined as:
\begin{equation}\label{eq:svr}
\mathrm{SVR}=0.6\,\mathrm{CS}+0.4\,\mathrm{OS}.
\end{equation}
\textbf{Parameter Match Rate (PMR)} measures the degree of alignment between generated and ground-truth parameters by verifying that both parameter names (e.g., InterArrivalTime, ProcessTime, SetupTime) and their assigned values including distributions and arguments are identical. Any discrepancy, such as \texttt{exponential(10)} versus \texttt{exponential(15)}, is treated as a mismatch. Let Pmatch denote the number of parameters whose name, type, and value all match, and Ptotal represent the total number of expected parameters. The metric is then computed as:
\begin{equation}\label{eq:pmr}
\mathrm{PMR}=\frac{P_{\text{match}}}{P_{\text{total}}}.
\end{equation}
\textbf{Execution Success Rate (ESR)} evaluates the executability of generated scripts within FlexSim. It measures whether each script can be imported and executed successfully without manual correction or runtime failure. Let $S_{\text{success}}$ denote the number of scripts that compile and run correctly, and $S_{\text{total}}$ represent the total number of evaluated samples. The metric is then defined as:
\begin{equation}\label{eq:esr}
\mathrm{ESR}=\frac{S_{\text{success}}}{S_{\text{total}}}.
\end{equation}
Although BLEU-4 fails to reflect structural or functional accuracy, it is included as a supplementary metric to quantify surface-level textual overlap and enable comparison with prior text-based studies.

%% file: sec/5_experiments.tex
\section{Experiments}
\subsection{Experimental Setup}

All experiments were conducted on a server equipped with eight NVIDIA L40 GPUs. Our dataset consists of 120,285 prompt–sketch–code triplets, which were randomly split into 90\% training, 5\% validation, and 5\% test sets. Each model is trained for 10 epochs, with all settings kept consistent across experiments. We report results on the test set. 

\subsection{LLM Evaluation}
\label{sec:llm_evaluation}
\begin{figure}[!htbp]
    \centering
    \includegraphics[width=1\linewidth]{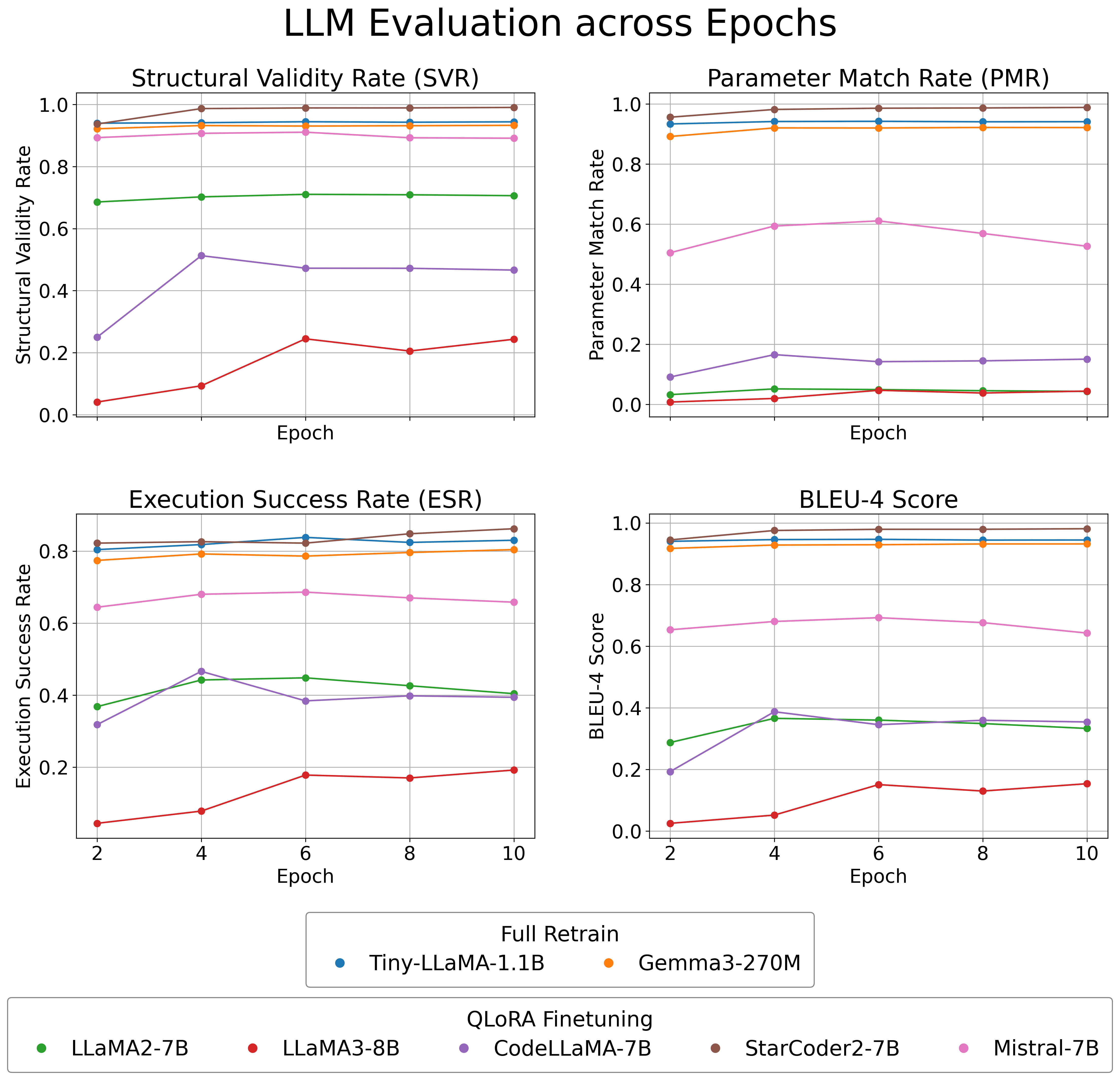}
    \caption{LLM Evaluation across Epochs. 
    Performance of seven LLM baselines over 10 epochs on four metrics.}
    \label{fig:llm_curves}
    \vspace*{-5pt}
\end{figure}

\begin{figure}[!htbp]
  \centering
  \includegraphics[width=\linewidth]{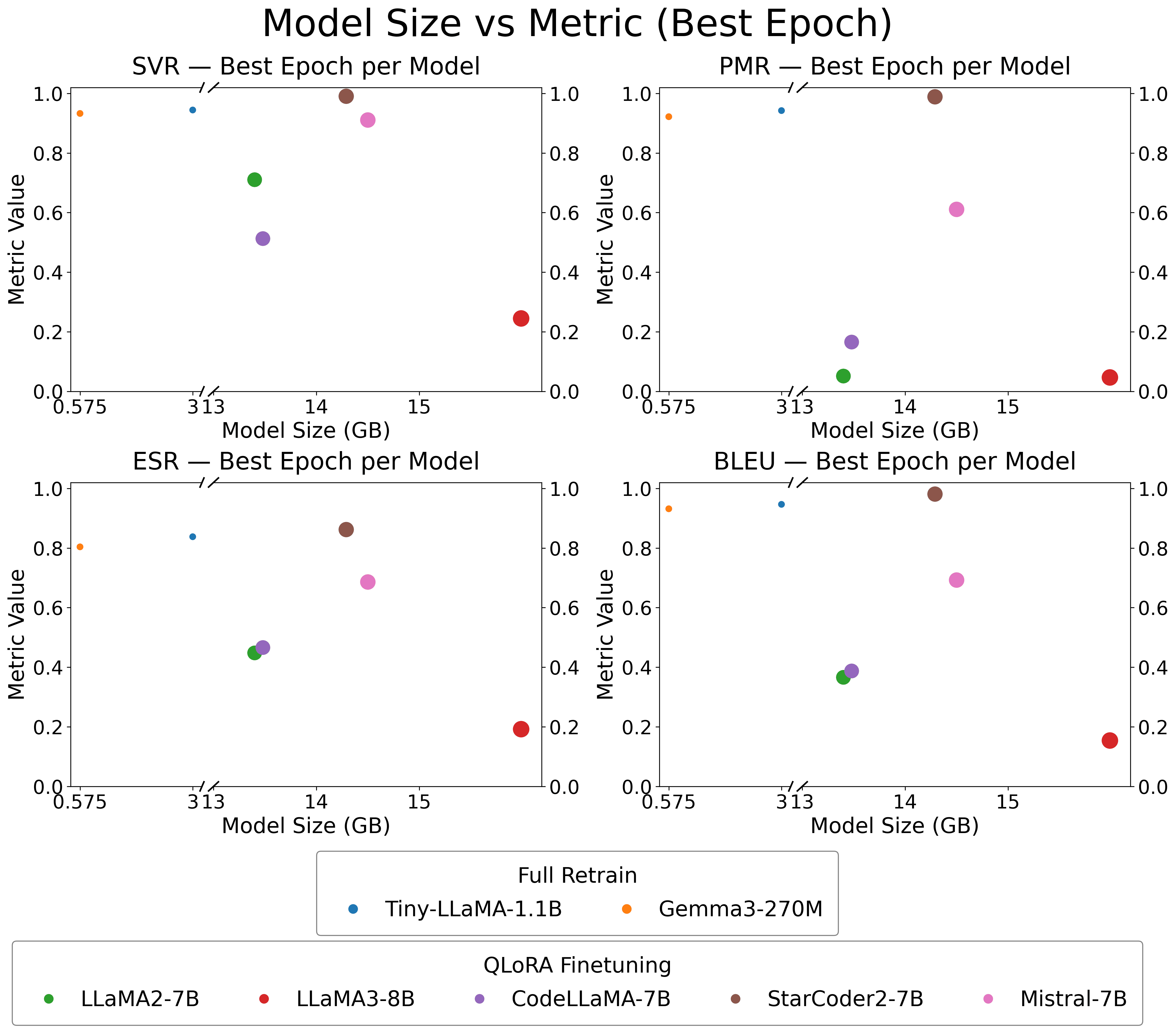}
  \caption{Model size vs. best epoch performance on four metrics.}
  \label{fig:model_size_metrics}
  \vspace*{-5pt}
\end{figure}

\begin{figure*}[t]
    \centering
    \includegraphics[width=\textwidth]{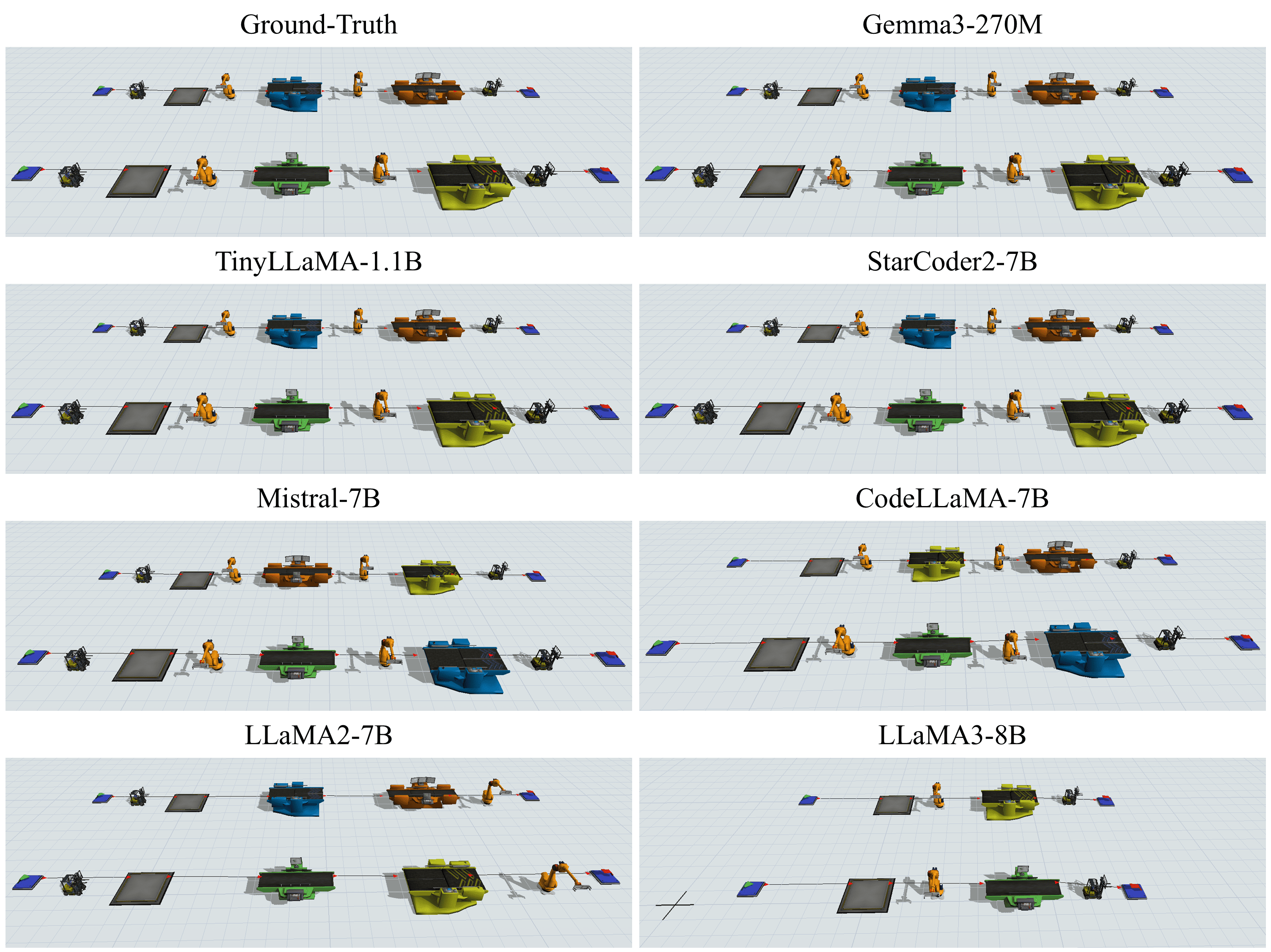}
    \caption{Qualitative comparison on a dual-line prompt. The three top-scoring backbones reported in Section~\ref{sec:llm_evaluation} closely reproduce the ground-truth layout. The remaining models show structural deviations consistent with their lower quantitative scores.}
    \label{fig:5.2Quant}
    \vspace*{-10pt}
\end{figure*}

We compare seven open-source LLM backbones on our GDT evaluation metrics (as shown in Figure~\ref{fig:llm_curves}). Figure~\ref{fig:model_size_metrics} indicates that code-pretrained StarCoder2-7B and fully retrained compact backbones, TinyLLaMA-1.1B and Gemma3-270M, outperform larger general-purpose LLaMA variants, which confirms that scale alone is insufficient for FlexScript generation. A qualitative example in Figure~\ref{fig:5.2Quant} mirrors this pattern, with the three top-scoring backbones recovering the ground-truth layout while the others show structural deviations. StarCoder2-7B attains the best scores across all metrics, converges swiftly, and remains dominant throughout training as shown in Figure~\ref{fig:llm_curves}. This advantage is consistent with its extensive pretraining on code, which aligns well with the structural demands of FlexScript. TinyLLaMA-1.1B ranks second overall, demonstrating that small backbones can become competitive when fully retrained on our dataset. Gemma3-270M, also fully retrained, achieves solid results yet trails TinyLLaMA-1.1B, suggesting that extremely small parameter counts still cap representational capacity. In contrast, general-purpose models do not match the above backbones, likely because their pretraining emphasizes broad natural language rather than structured simulation code. Guided by single-modality performance (as shown in Table~\ref{tab:llm_baselines}) and stable learning curves, we therefore select StarCoder2-7B and TinyLLaMA-1.1B as backbones for our multimodal extensions in Section~\ref{sec:Ablation Study}.

\begin{table}[!htbp]
\centering
\caption{LLM baselines on FlexScript. Each model’s best score}
\label{tab:llm_baselines}
\setlength{\tabcolsep}{4pt}
\resizebox{\columnwidth}{!}{%
\begin{tabular}{lcccc}
\toprule
Model & Best SVR & Best PMR & Best ESR & Best BLEU-4 \\
\midrule
Gemma3-270M     & 0.9328  & 0.9219  & 0.8040  & 0.9318\\
TinyLLaMA-1.1B  & 0.9444  & 0.9424  & 0.8380  & 0.9467\\
Mistral-7B      & 0.9107  & 0.6108  & 0.6860  & 0.6925\\
LLaMA2-7B       & 0.7104  & 0.0513  & 0.4480  & 0.3660\\
CodeLLaMA-7B    & 0.5127  & 0.1654  & 0.4660  & 0.3874\\
StarCoder2-7B   & \textbf{0.9905} & \textbf{0.9886} & \textbf{0.8620}  & \textbf{0.9811}\\
LLaMA3-8B       & 0.2447  & 0.0466  & 0.1920 & 0.1539\\
\bottomrule
\end{tabular}%
}
\vspace*{-5pt}
\end{table}

\begin{figure}[!htbp]
  \centering
  \includegraphics[width=\linewidth]{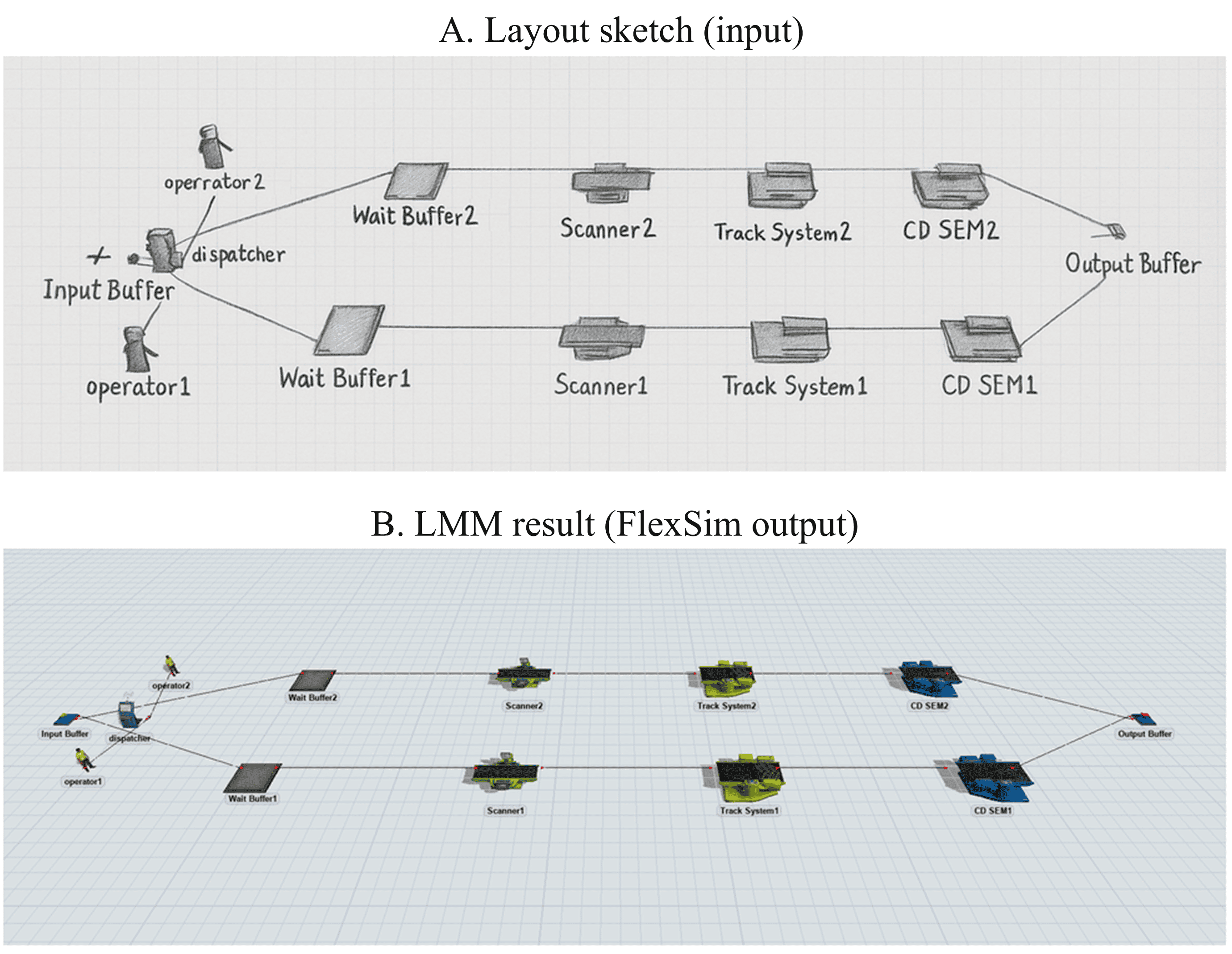}
  \caption{Qualitative effect of multimodal coupling.
  Preserving two-line topology
  and object ordering, consistent with score gains in Table~\ref{tab:ablation_tiny} and Table~\ref{tab:ablation_starcoder}.}
  \label{fig:qual_mm_vertical}
\end{figure}

\subsection{Ablation Study}
\label{sec:Ablation Study}
We perform an ablation study on multimodal integration strategies by evaluating two representative backbones, TinyLLaMA-1.1B and StarCoder2-7B, in combination with vision encoders and connector modules. 

\begin{table}[!htbp]
\centering
\setlength{\tabcolsep}{3pt}
\renewcommand{\arraystretch}{1.05}
\caption{Ablation study of TinyLLaMA-1.1B with vision encoders and connector modules. 
The first row is the text-only baseline.}
\label{tab:ablation_tiny}
\resizebox{\columnwidth}{!}{%
\begin{tabular}{llcccc}
\toprule
Vision Encoder & Connector & SVR & PMR & ESR & BLEU-4 \\
\midrule
\multicolumn{2}{l}{TinyLLaMA-1.1B (LLM-only)} & \textbf{0.9444} & 0.9424 & 0.8380 & 0.9467 \\
CLIP & Linear Projection      & 0.8911 & 0.9284 & 0.8020 & 0.9164 \\
CLIP & Perceiver-style Resampler    & 0.8607 & 0.9311 & 0.8080 & 0.9205 \\
CLIP & Q-Former                & 0.8535 & 0.9311 & 0.7960 & 0.9150 \\
CLIP & Two-Layer MLP          & 0.9059 & 0.9229 & 0.8300 & 0.9238 \\
OpenCLIP & Linear Projection  & 0.9408 & \textbf{0.9505} & \textbf{0.8820} & \textbf{0.9482} \\
OpenCLIP & Perceiver-style Resampler& 0.9144 & 0.9330 & 0.8040 & 0.9204 \\
OpenCLIP & Q-Former            & 0.9314 & 0.9422 & 0.8500 & 0.9265 \\
OpenCLIP & Two-Layer MLP      & 0.9243 & 0.9403 & 0.8220 & 0.9222 \\
\bottomrule
\end{tabular}%
}
\vspace*{-3pt}
\end{table}

\begin{table}[!htbp]
\centering
\setlength{\tabcolsep}{3pt}
\renewcommand{\arraystretch}{1.05}
\caption{Ablation study of StarCoder2-7B with vision encoders and connector modules. 
The first row is the text-only baseline.}
\label{tab:ablation_starcoder}
\resizebox{\columnwidth}{!}{%
\begin{tabular}{llcccc}
\toprule
Vision Encoder & Connector & SVR & PMR & ESR & BLEU-4 \\
\midrule
\multicolumn{2}{l}{StarCoder2-7B (LLM-only)} & 0.9905 & 0.9886 & 0.8620 & 0.9811 \\
CLIP & Linear Projection       & 0.9958 & 0.9930 & 0.8640 & 0.9874 \\
CLIP & Perceiver-style Resampler     & 0.9903 & \textbf{0.9928} & 0.8480 & 0.9843 \\
CLIP & Q-Former                 & 0.9825 & 0.9876 & 0.8380 & 0.9724 \\
CLIP & Two-Layer MLP           & 0.9861 & 0.9932 & 0.8420 & 0.9829 \\
OpenCLIP & Linear Projection   & 0.9958 & 0.9857 & 0.8720 & 0.9866 \\
OpenCLIP & Perceiver-style Resampler & 0.9857 & 0.9849 & 0.8600 & 0.9862 \\
OpenCLIP & Q-Former             & 0.9948 & 0.9913 & 0.8660 & 0.9868 \\
OpenCLIP & Two-Layer MLP       & \textbf{0.9990} & 0.9922 & \textbf{0.8740} & \textbf{0.9886} \\
\bottomrule
\end{tabular}%
}
\vspace*{-3pt}
\end{table}

\begin{figure*}[t]
  \centering
  \includegraphics[width=\linewidth]{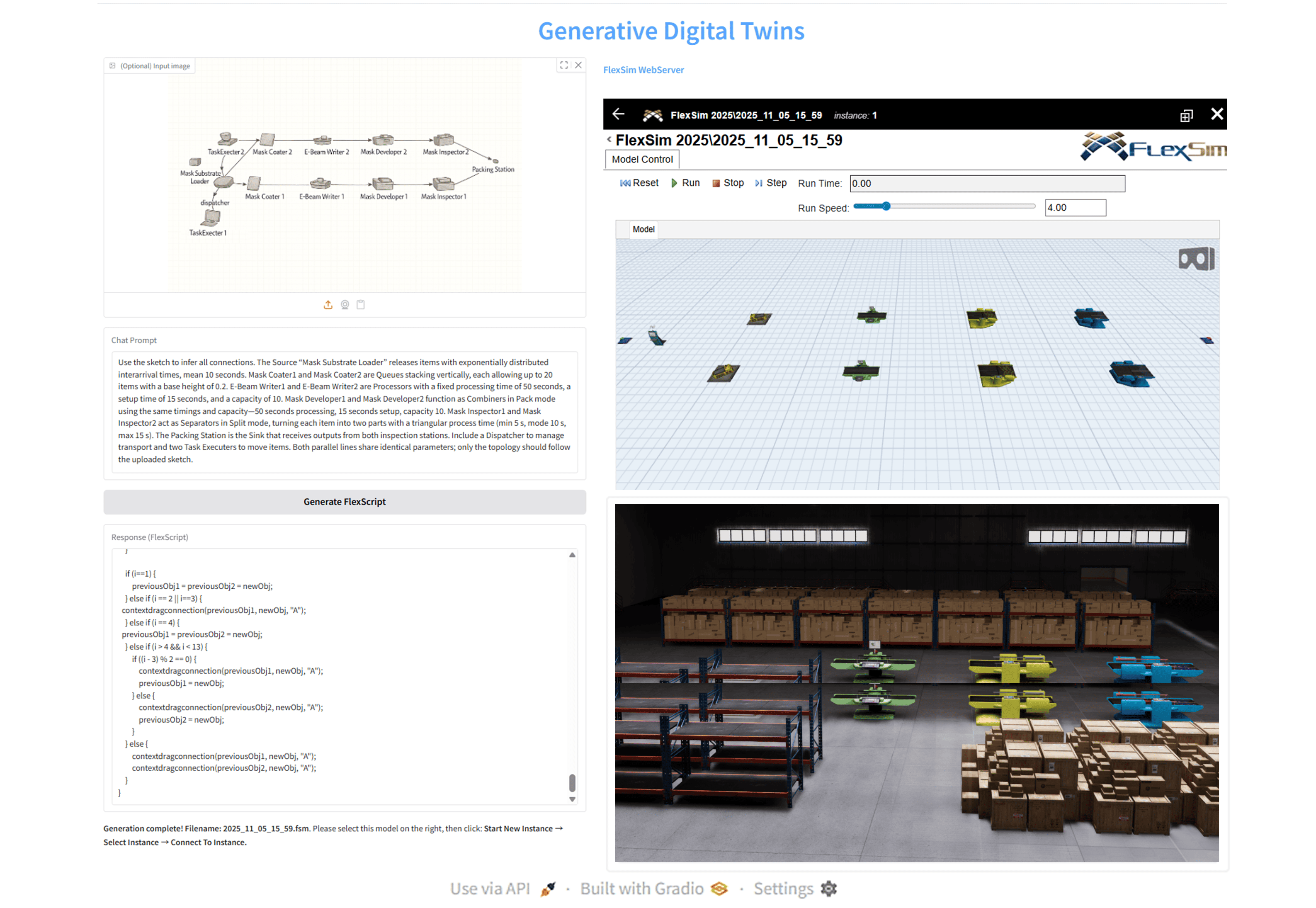}
  \caption{User interface of GDT. The left panel accepts a draft layout image and a prompt, the model generates FlexScript automatically. The right panels visualize the instantiated 3D scene and results rendered in NVIDIA Omniverse, enabling quick validation before execution.}
  \label{fig:gdt-ui}
  \vspace*{-10pt}
\end{figure*}

As shown in Table~\ref{tab:ablation_tiny}, for TinyLLaMA-1.1B, OpenCLIP + Linear Projection achieves the best balance, increasing PMR from 0.9424 to 0.9505 and ESR from 0.8380 to 0.8820, while slightly improving BLEU-4 to 0.9482. This indicates that visual integration can substantially enhance the robustness of execution. 

As shown in Table~\ref{tab:ablation_starcoder}, for StarCoder2-7B, the LLM-only baseline already achieves near-ceiling text-only results. A qualitative example in Figure~\ref{fig:qual_mm_vertical} mirrors this pattern, showing faithful two-line topology and correct station order under multimodal coupling. Nonetheless, multimodal integration brings further improvements in execution robustness, OpenCLIP with Two-Layer MLP achieves the highest SVR (0.9990), ESR (0.8740), and BLEU-4 (0.9886). The multimodal variants significantly strengthen structural validity and execution feasibility.

In summary, OpenCLIP consistently outperforms CLIP across both backbones, while lightweight connectors (Linear Projection, Two-Layer MLP) remain competitive with Q-Former. 
TinyLLaMA benefits more from multimodal integration due to its smaller scale, whereas StarCoder2 maintains strong text-only ability but still improves in SVR, ESR and BLEU-4 when augmented with vision.

\subsection{Final Model: VLSM}
\label{sec:Final Model}
Based on the ablation results in Section~\ref{sec:Ablation Study}, we present two multimodal models optimized for FlexScript generation. For TinyLLaMA-1.1B, we adopt OpenCLIP as the vision encoder with a Linear Projection connector. This pairing provides the best balance across the evaluation metrics. This configuration is denoted as \textbf{VLSM-1.1B}.  
For StarCoder2-7B, we adopt the OpenCLIP encoder with Two-Layer MLP, which achieves near-perfect SVR and the highest ESR. This configuration is denoted as \textbf{VLSM-7B}.  

Together, these models constitute the VLSM family. To our knowledge, they are the first multimodal models specifically optimized for FlexScript generation, combining domain-adapted LLM backbones with vision-grounded connectors. VLSM demonstrates strong structural accuracy, robust parameter alignment, and reliable executability, providing a practical foundation for simulation-driven digital-twin automation. A prototype of the full workflow is shown in Figure~\ref{fig:gdt-ui}.

%% file: sec/6_conclusion.tex
\section{Conclusion}
We present the GDT framework, unifying multimodal code generation and evaluation. We introduce a 120K-scale domain dataset and three evaluation metrics that assess structural validity, parameter consistency, and execution feasibility. Our VLSM family couples visual perception with program synthesis and delivers robust prompt-to-code generation in FlexSim. Experiments and ablations show that code-pretrained backbones with visual conditioning improve spatial alignment and execution robustness, leading to consistent gains across evaluation metrics. Together, the framework, data, and metrics establish a reproducible benchmark and a solid foundation for industrial simulation automation, replacing labor-intensive digital-twin construction with reliable layout-aware FlexScript generation.

%% file: sec/X_suppl.tex
\clearpage
\setcounter{page}{1}
\maketitlesupplementary

\appendix

\section*{A. Reproducibility and Code Release}

We are committed to research reproducibility and transparency. Accordingly, we have released the source code for our proposed method, covering model definitions, configuration settings, and training scripts necessary to reproduce the main experiments. Comprehensive resources, including the full dataset and pretrained model checkpoints, are being prepared for public release and will be accessible via the same repository.

\noindent\textbf{GitHub:} \\
\url{https://github.com/danielhsu2014/GDT-VLSM}

% ======================= B. Discussion =======================
\section*{B. Discussion}

\subsection*{B.1. Limitations}
The proposed Vision-Language Simulation Models generate executable FlexScript for factory layouts within the scope of GDT-120K. Transfer to other simulators or unseen production logic is not guaranteed and code patterns learned from FlexSim classes may not carry over. TinyLLaMA-1.1B focuses on in-domain learning and can be weaker for broad common-sense reasoning. StarCoder2-7B has stronger priors on code structure but requires higher compute and memory demand even with four-bit quantization. The current system is single-turn from prompt to code without interactive repair so users still need verification and debugging. Compatibility with future FlexScript or API changes is not ensured and would require adaptation. Evaluation emphasizes structural and executable correctness while satisfaction of high-level intent is not explicitly measured and remains open.

\subsection*{B.2. Societal Impact}
Authoring digital twins in FlexSim is labor-intensive because it requires manual object placement, parameter configuration, and logic scripting, which slows adoption and does not scale. Our VLSM reframes this workflow by letting users specify layouts with sketches and prompts while the model synthesizes executable FlexScript, which reduces manual scripting time and lowers the barrier for non-experts. 

The GDT-120K corpus and the task-specific metrics SVR, PMR, and ESR promote measurable quality control and reproducibility for simulation code. Experiments in the main paper report near-perfect structural accuracy and high execution robustness, which supports faster prototyping, more reliable planning, and fewer rework cycles in digital-twin projects. Together these elements form a practical foundation that replaces labor-intensive construction with layout-aware code generation and enables scalable deployment in manufacturing. 

Taken together, these capabilities open a new chapter for digital twins. Layout-aware code generation with standardized evaluation and reproducible training turns descriptive intent into executable systems, shortening time-to-validation and enabling broader participation by non-experts. We anticipate gains in design exploration, safety analysis, and sustainable operations as simulation becomes a routine, verifiable step in manufacturing practice.

% ======================= C. Preliminaries =======================
\section*{C. Preliminaries}
We outline the language backbones and the vision-language alignment modules evaluated in this work. The main backbones are StarCoder2-7B and TinyLLaMA-1.1B. StarCoder2-7B is a code-oriented model adapted with parameter-efficient updates and serves as our strongest backbone for GDT. TinyLLaMA-1.1B is a compact model trained with full parameter updates in-domain and offers a small footprint with competitive accuracy.

We also benchmark additional backbones under the same decoding policy to form language-only baselines. Gemma3-270M is a small model trained in-domain and provides a lightweight reference. Mistral-7B is a general-purpose model adapted to the task and tests transfer from broad pretraining. LLaMA2-7B is another general backbone used for comparison on the same data and prompts. CodeLLaMA-7B is a code-focused LLaMA variant that probes the benefit of code priors beyond StarCoder2. LLaMA3-8B is a recent general model included to measure progress on the base task.

% ======================= D. Implementation Details =======================
\section*{D. Implementation Details}

Data come from GDT-120K with train, validation, and test splits. During inference, the model receives an instruction prefix and a prompt, then decodes until the end of sequence. We evaluate SVR, PMR, ESR, and BLEU as defined in the main paper. SVR checks object declarations and the connection graph, and PMR verifies the required parameters. ESR loads each script in FlexSim and executes a short run. BLEU is reported for reference and is not the primary indicator of correctness.

\subsection*{D.1. Model Settings}
We first present the language-only setups to isolate backbone behavior. The finetuning hyperparameters that apply to StarCoder2-7B, Mistral-7B, LLaMA2-7B, CodeLLaMA-7B, and LLaMA3-8B are summarized in Table~\ref{tab:ft_single}. The full-retraining hyperparameters that apply to TinyLLaMA-1.1B, Gemma3-270M, and compatible small backbones are summarized in  Table~\ref{tab:fr_single}. These settings define tokenization, sequence length, label masking, learning-rate schedule, and batching and they serve as the base configuration.

% ==== Single-LLM Hyperparameters: Finetuning (QLoRA) ====
\begin{table}[!htbp]
\centering
\caption{Single LLM hyperparameters for finetuning with QLoRA}
\label{tab:ft_single}
\setlength{\tabcolsep}{4pt}
\resizebox{\columnwidth}{!}{%
\begin{tabular}{ll}
\toprule
Field & Value \\
\midrule
Tokenizer & AutoTokenizer family with pad token equal to EOS \\
Instruction prefix & ``Give me a FlexScript.'' \\
Sequence length & 4096 \\
Label masking & ignore prompt tokens with -100 \\
Quantization & 4-bit NF4 with double quant \\
LoRA rank    & 32 \\
LoRA alpha   & 64 \\
LoRA dropout & 0.05 \\
LoRA targets & attention and feed-forward projections \\
Batching & per-device train batch: 8, accumulation: 4 \\
Epochs & 10 \\
Learning rate & $1\times10^{-4}$ constant with warmup steps 10 \\
Precision and GC & fp16, no gradient checkpointing \\
DDP flag & find unused parameters false \\
\bottomrule
\end{tabular}%
}
\end{table}

% ==== Single-LLM Hyperparameters: Full Retraining ====
\begin{table}[!htbp]
\centering
\caption{Single LLM hyperparameters for full retraining}
\label{tab:fr_single}
\setlength{\tabcolsep}{4pt}
\resizebox{\columnwidth}{!}{%
\begin{tabular}{ll}
\toprule
Field & Value \\
\midrule
Tokenizer & AutoTokenizer family with pad token equal to EOS \\
Instruction prefix & ``Give me a FlexScript.'' \\
Sequence length & 4096 \\
Label masking & ignore prompt tokens with -100 \\
Batching & per-device train batch: 2, accumulation: 8 \\
Epochs & 10 \\
Learning rate & $1\times10^{-4}$ cosine with warmup steps 100 \\
Precision and GC & fp16 and gradient checkpointing \\

\bottomrule
\end{tabular}%
}
\end{table}

We next introduce the multimodal instantiations that pair OpenCLIP with TinyLLaMA or with StarCoder2. The connector architectures and trainable blocks are summarized in Table~\ref{tab:conn_tiny} and Table~\ref{tab:conn_star}.

\begin{table}[!htbp]
\centering
\caption{Connector specification for OpenCLIP with TinyLLaMA}
\label{tab:conn_tiny}
\setlength{\tabcolsep}{4pt}
\resizebox{\columnwidth}{!}{%
\begin{tabular}{ll}
\toprule
Field & Value \\
\midrule
Connector and backbone & OpenCLIP with TinyLLaMA \\
Connector architecture & single linear projection \\
Trainable blocks & OpenCLIP and linear projector \\
LLM quantization and adapter & 4-bit NF4, bf16 compute and LLM frozen \\
Image token labels & set to -100 \\
Max sequence length & 4096 \\
Notes & efficient with limited nonlinearity \\
\bottomrule
\end{tabular}%
}
\end{table}

\begin{table}[!htbp]
\centering
\caption{Connector specification for OpenCLIP with StarCoder2}
\label{tab:conn_star}
\setlength{\tabcolsep}{4pt}
\resizebox{\columnwidth}{!}{%
\begin{tabular}{ll}
\toprule
Field & Value \\
\midrule
Connector and backbone & OpenCLIP with StarCoder2 \\
Connector architecture & two-layer perceptron with GELU \\
Trainable blocks & OpenCLIP and two layer projector and LoRA loaded \\
LLM quantization and adapter & 4-bit NF4, bf16 compute and LoRA loaded \\
Image token labels & set to -100 \\
Max sequence length & 4096 \\
Notes & stronger alignment at higher resource cost \\
\bottomrule
\end{tabular}%
}
\end{table}

\subsection*{D.2. Training Details}
The optimization settings for the multimodal variants are summarized in Table~\ref{tab:train_star} and Table~\ref{tab:train_tiny}. Both implementations use consistent batching rules and select the checkpoint with the lowest validation loss.

\begin{table}[!htbp]
\centering
\caption{Training settings for the StarCoder2 variant}
\label{tab:train_star}
\setlength{\tabcolsep}{4pt}
\resizebox{\columnwidth}{!}{%
\begin{tabular}{ll}
\toprule
Field & Value \\
\midrule
Backbone and variant & StarCoder2-7B with two-layer connector \\
Update strategy & base frozen \\
Precision & 4-bit weights, bf16 compute \\
Base learning rate & $1\times10^{-5}$ \\
Schedule and warmup & cosine schedule with 50 warmup steps \\
Micro batch & 1 \\
Accumulation & 4 \\
Checkpoint selection & lowest validation loss \\
\bottomrule
\end{tabular}%
}
\end{table}

\begin{table}[!htbp]
\centering
\caption{Training settings for the TinyLLaMA variant}
\label{tab:train_tiny}
\setlength{\tabcolsep}{4pt}
\resizebox{\columnwidth}{!}{%
\begin{tabular}{ll}
\toprule
Field & Value \\
\midrule
Backbone and variant & TinyLLaMA-1.1B with linear connector \\
Update strategy & full updates on vision and connector with LLM frozen \\
Precision & 4-bit LLM with bf16 compute \\
Base learning rate & $1\times10^{-5}$ \\
Schedule and warmup & cosine schedule with 50 warmup steps \\
Micro batch & 2 \\
Accumulation & 4 \\
Checkpoint selection & lowest validation loss \\
\bottomrule
\end{tabular}%
}
\end{table}

% ======================= E. Additional Experiments =======================

\begin{figure}[!htbp]
\centering
\includegraphics[width=1\linewidth]{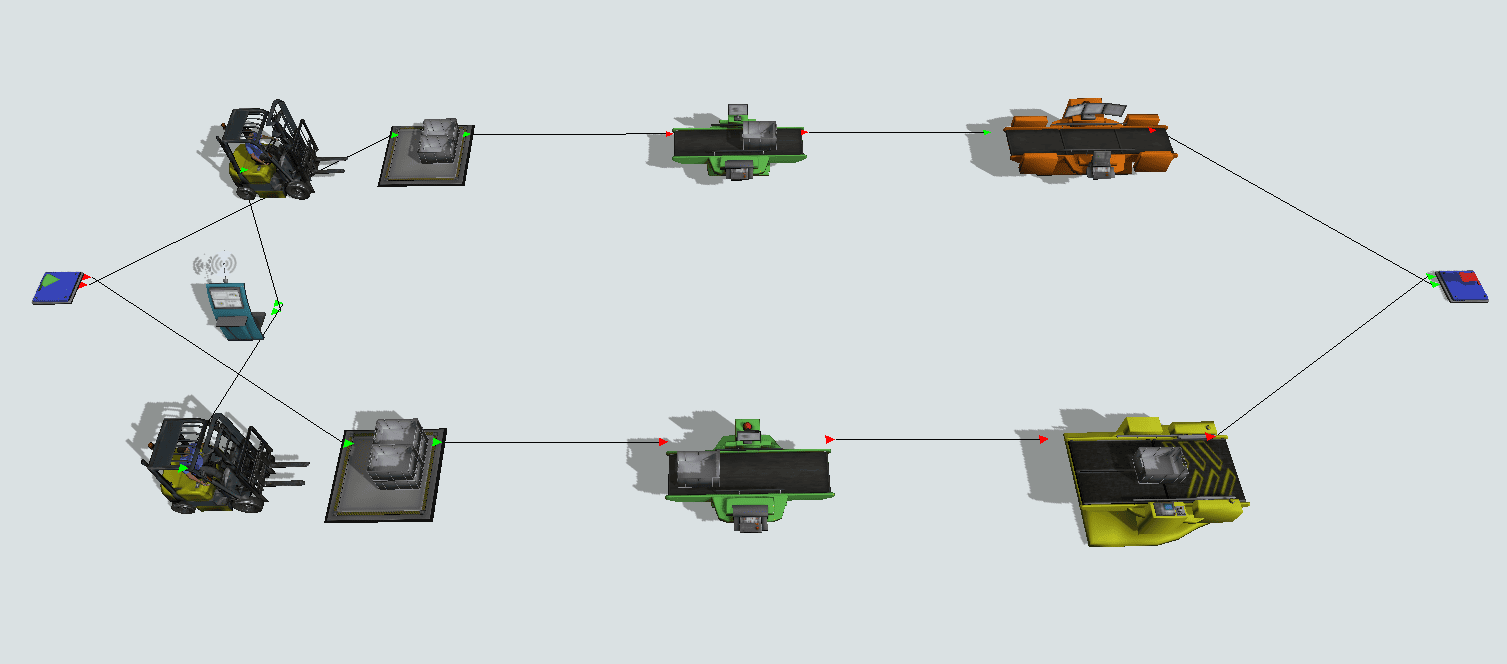}
\caption{Runtime snapshot of a generated simulation executing in FlexSim.}
\label{fig:flexsim_runtime}
\end{figure}

\begin{figure*}[t]
\vspace*{35pt}
\centering
\includegraphics[width=\linewidth]{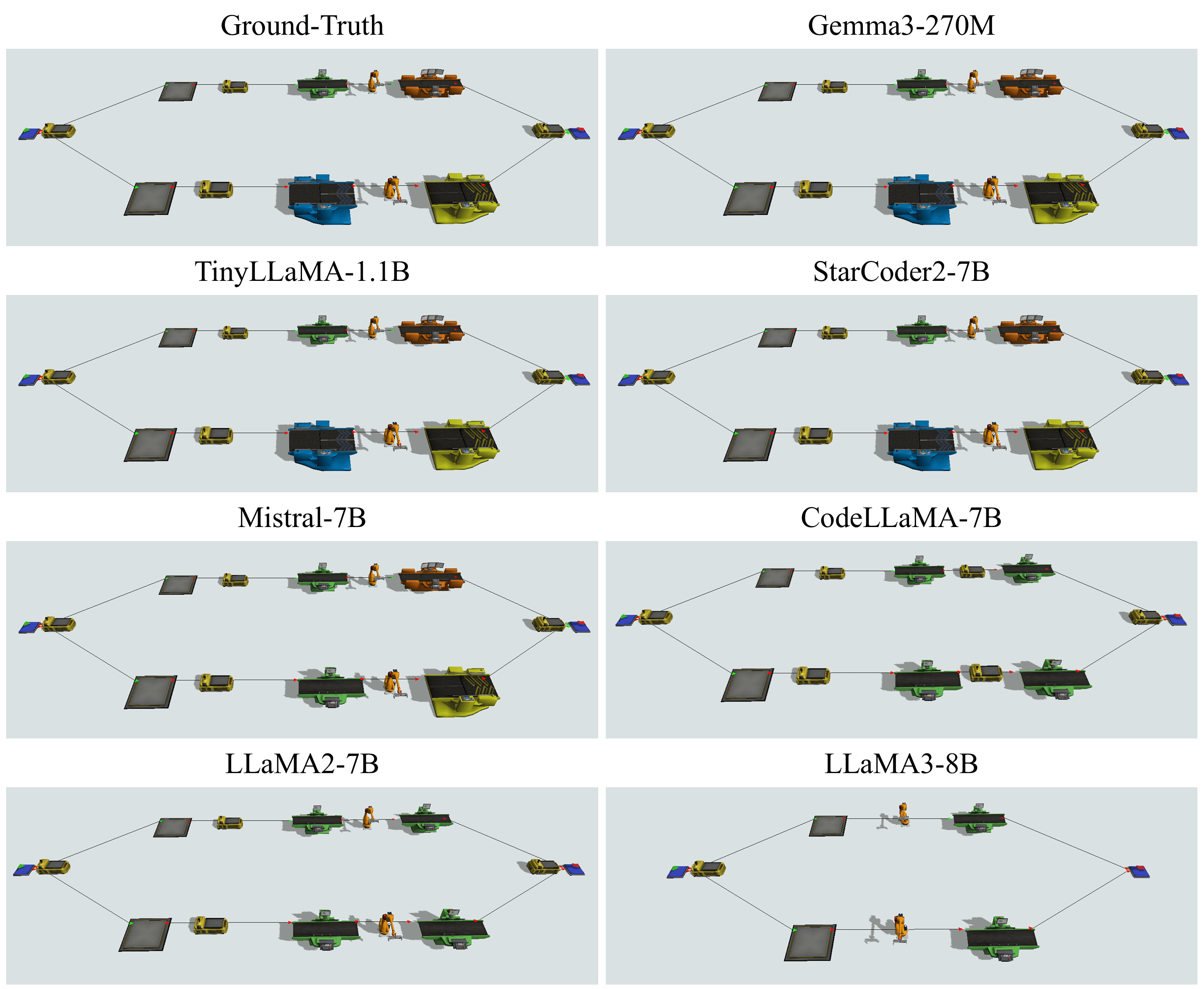}
\caption{Qualitative comparison of layouts generated by different language backbones against the Ground-Truth.}
\label{fig:qual1}
\vspace*{25pt}
\end{figure*}

\section*{E. Additional Experiments}
\begin{figure*}[t]
\centering
\includegraphics[width=0.9\linewidth]{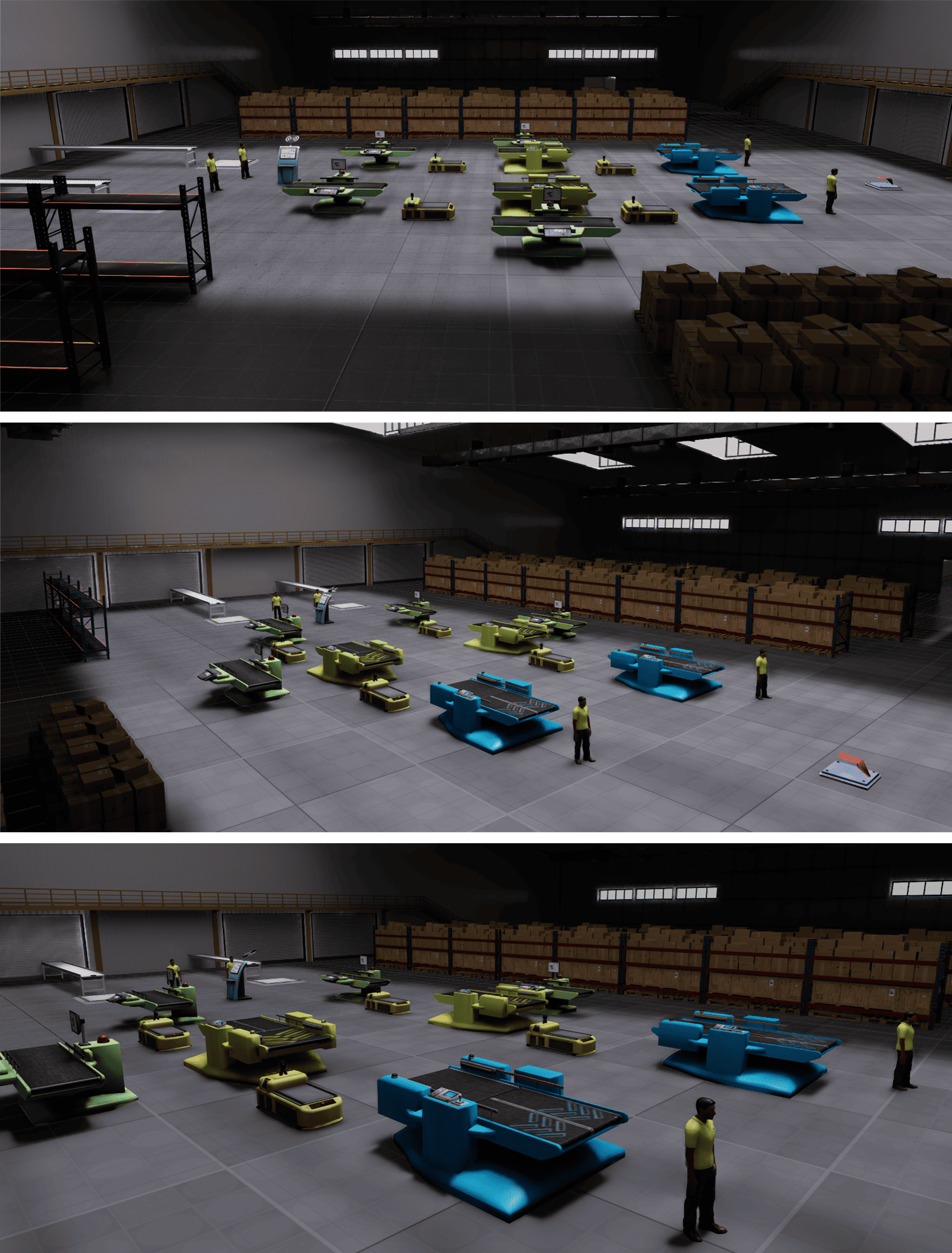}
\caption{Varying perspectives of a single factory layout generated by our VLSM-7B and rendered in NVIDIA Omniverse.}
\label{fig:qual2}
\end{figure*}

\begin{figure*}[t]
\centering
\includegraphics[width=0.9\linewidth]{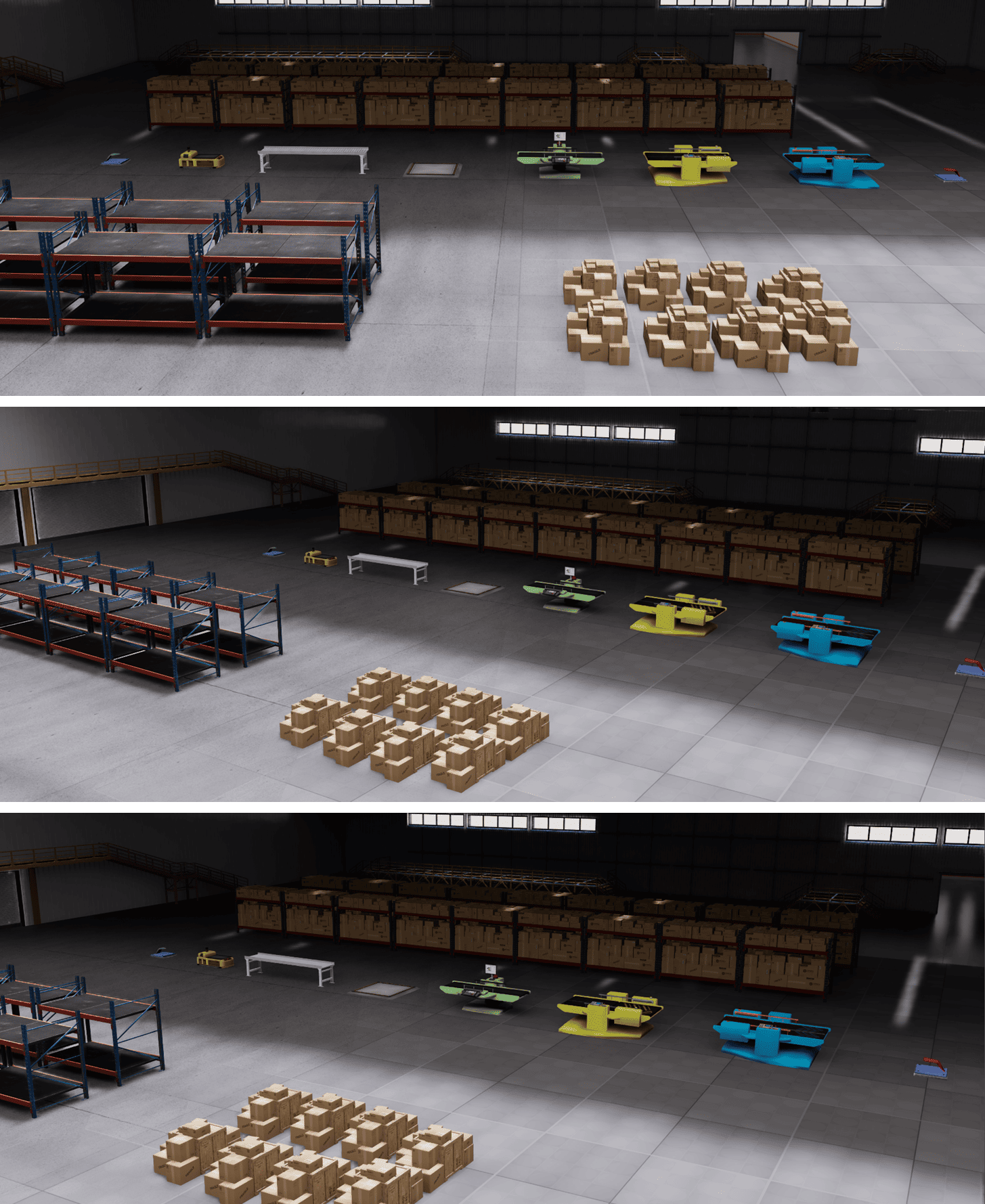}
\caption{Varying perspectives of a single factory layout generated by our VLSM-7B and rendered in NVIDIA Omniverse.}
\label{fig:qual3}
\end{figure*}

\begin{figure*}[t]
\centering
\includegraphics[width=0.9\linewidth]{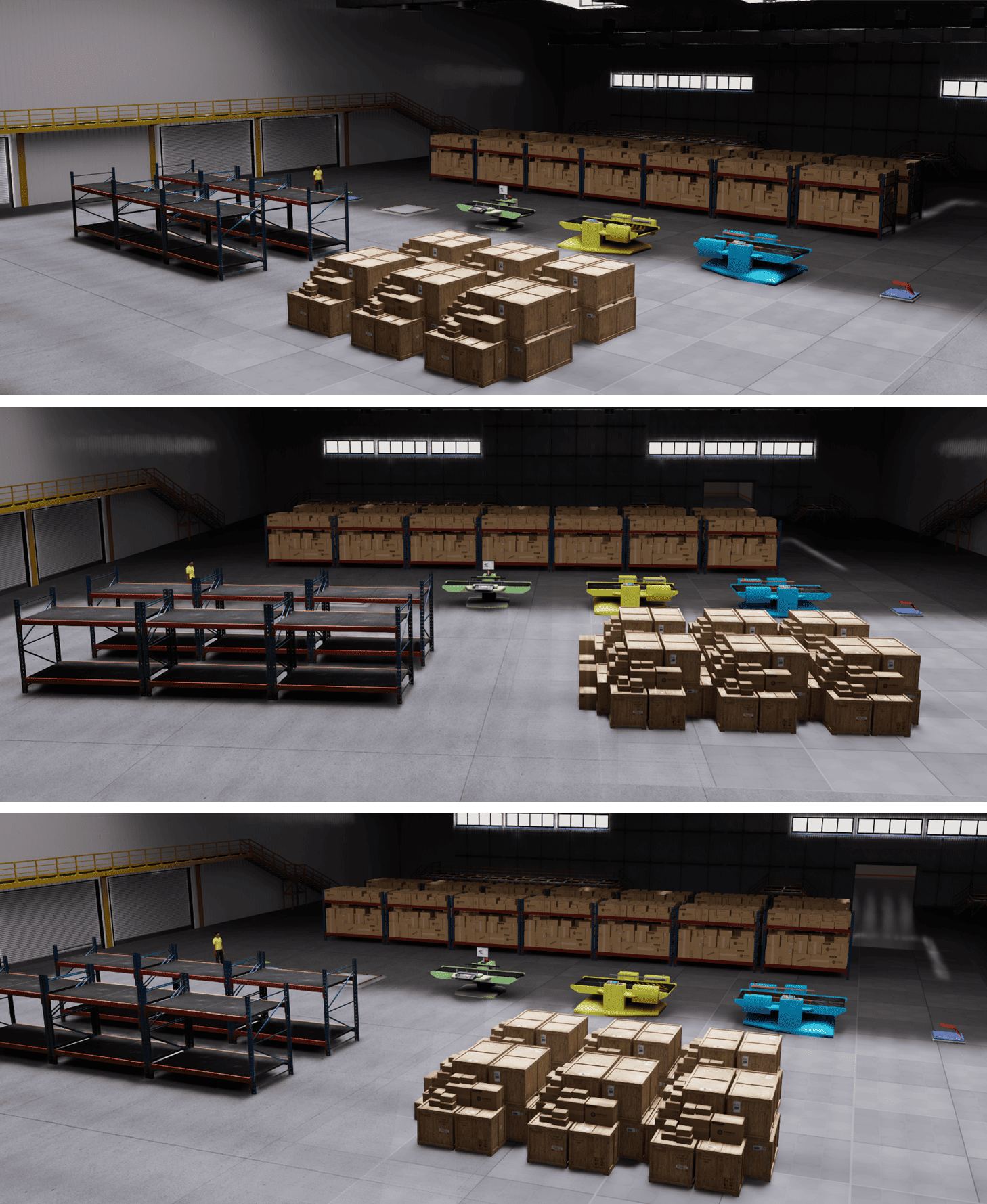}
\caption{Varying perspectives of a single factory layout generated by our VLSM-7B and rendered in NVIDIA Omniverse.}
\label{fig:qual4}
\end{figure*}

In this section, we present an extended qualitative analysis to further demonstrate the capability of our VLSM in handling complex industrial scenarios. Visual inspection of the executable simulations provides crucial insights into the structural coherence and logic validity of the generated code. Figure~\ref{fig:flexsim_runtime} captures the runtime behavior of the synthesized FlexScript, confirming that the generated layouts are not merely static arrangements but fully functional systems with active object interactions. The side-by-side comparison in Figure~\ref{fig:qual1} visually evidences that while general-purpose LLMs often struggle with spatial connectivity, code-specialized models closely reconstruct the ground-truth topology. Finally, we highlight the generative quality of our best-performing VLSM-7B backbone using high-fidelity renderings in NVIDIA Omniverse. Figure~\ref{fig:qual2} demonstrates the structural consistency of a complex factory layout across multiple viewing angles. Figure~\ref{fig:qual3} and Figure~\ref{fig:qual4} display the model's versatility in synthesizing diverse, large-scale warehousing environments, including bulk storage and sorting areas, respectively.

